\documentclass[runningheads]{llncs}
\usepackage{cite}
\usepackage{amsmath,amssymb,amsfonts}
\usepackage{graphicx}
\usepackage{algorithm}
\usepackage{algorithm,algpseudocode}

\usepackage{textcomp}
\usepackage{xcolor}
\usepackage[labelsep=quad,indention=10pt]{subfig}

\begin{document}
\title{Deep Visual Anomaly detection with Negative Learning}
%
%
\author{Jin-Ha Lee\inst{1,2}\orcidID{0000-1111-2222-3333} \and
Marcella Astrid\inst{1,2}\orcidID{0000-0003-1432-6661} \and
Muhammad Zaigham Zaheer\inst{1,2}\orcidID{0000-0001-8272-1351} \and
Seung-Ik Lee\inst{1,2}\orcidID{0000-0003-2986-7540}}
\authorrunning{J.H.Lee et al.}
%
\institute{University of Science and Technology (UST), 217, Gajeong-ro, Yuseong-gu, Daejeon, Republic of Korea\\
\email{\{jhlee, marcella.astrid, mzz\}@ust.ac.kr}\\
\and
Electronics and Telecommunications Research Institute (ETRI), 218, Gajeong-ro, Yuseong-gu, Daejeon, Republic of Korea\\
\email{\{the\_silee\}@etri.re.kr}}
\maketitle              
\begin{abstract}
With the increase in the learning capability of deep convolution-based architectures, various applications of such models have been proposed over time.
In the field of anomaly detection, improvements in deep learning opened new prospects of exploration for the researchers whom tried to automate the labor-intensive features of data collection. 
First, in terms of data collection, it is impossible to anticipate all the anomalies that might exist in a given environment.
Second, assuming we limit the possibilities of anomalies, it will still be hard to record all these scenarios for the sake of training a model
Third, even if we manage to record a significant amount of abnormal data, it's laborious to annotate this data on pixel or even frame level.
Various approaches address the problem by proposing one-class classification using generative models trained on only normal data.
In such methods, only the normal data is used, which is abundantly available and doesn't require significant human input.
However, such approaches have two drawbacks.
First, these are trained with only normal data and at the test time, given abnormal data as input, still generate normal-looking output. 
This happens due to the hallucination characteristic of generative models, which is not desirable in anomaly detection systems because of their need to be accurate and reliable. 
Next, these systems are not capable of utilizing abnormal examples, however small in number, during the training.  
In this paper, we propose anomaly detection with negative learning (ADNL), which employs the negative learning concept for the enhancement of anomaly detection by utilizing a very small number of labeled anomaly data as compared with the normal data during training.
The idea, which is fairly simple yet effective, is to limit the reconstruction capability of a generative model using the given anomaly examples.
During the training, normal data is learned as would have been in a conventional method, but the abnormal data is utilized to maximize loss of the network on abnormality distribution.
with this simple tweaking, the network not only learns to reconstruct normal data but also encloses the normal distribution far from the possible distribution of anomalies.
In order to evaluate the efficiency of our proposed method, we defined the baseline using Adversarial Auto-Encoder(AAE).
Our experiments show significant improvement in area under the curve(AUC) over conventional AAE.
An extensive evaluation, which has been carried out on the MNIST dataset as well as a locally recorded pedestrian dataset, is reported in this paper.

\end{abstract}
\section{Introduction}
Anomaly Detection is one of the most challenging tasks in the field of machine-learning which naturally makes it particularly interesting for researchers\cite{chandola2009anomaly, chalapathy2019deep}. 
Various variants of anomaly detection algorithms are being used in the field of signal processing\cite{lu2009network, ranney2006hyperspectral}, medical diagnosis\cite{stafford1994application}, network intrusion detection \cite{liu2008isolation,mukkamala2002intrusion, ryan1998intrusion}, and video surveillance\cite{mahadevan2010anomaly, kratz2009anomaly}.
Anomaly detection systems, although difficult to train, are expected to have high efficiency as such models are expected to reduce human labor for their corresponding domains.
Specifically, in video surveillance for the purpose of safety and security, autonomous anomaly detection systems contribute significantly by analysing hundreds of concurrent video streams which otherwise would take a lot of human resources as well as attention.
With the recent development in deep-learning algorithms it is becoming handy to to apply these low computational yet highly efficient algorithms for the learning problems.
2D and 3d convolutions, auto-encoders, object detectors, aaaaa  are few among many other algorithms which are being used for anomaly detection.

These techniques can be widely divided into two categories.
The first type is where the network is learnt based on normal as well as abnormal examples.
During the test time, each scenario is compared against the train model to identify anomalies. 
In contrary, the other type is the one in which learning is done using only normal data to define a corresponding distribution.
In order to check whether the test data is anomalous or not, it is compared against the learned normal distribution and checked if it is an outlier.  
However, each of these categories have its own limits or drawbacks.
For the formal category, one common problem is the absolute necessity of defining abnormality.
However, in real world scenarios anomalies cannot be anticipated.
There is always a possibility of something new happening and, if it is not learnt as abnormal during the training time, it might get overlooked during the test time.
Moreover, another problem that such system face is the lack of data.
It is mostly due to the limiting factor of defining the abnormalities.
Even if the actors are used to generate anomalous scenes, it is phisicaly impossible to create all possible scenarios.
In addition, this category of anomaly detection solutions require huge amount of labeled data for training which is laborious to obtain.

The rapid development in recent deep learning filed seems to meet those expectations, but still, some problems have remained.
One is, how to define Abnormal.
The definition of abnormal can be made by the definition of normal, which can be everything that is not normal.
The definition of normal is also difficult.
For example, when a car in on the highway, it is normal.
However, if the car is on the pedestrian road, that is abnormal.
Considering the situations, how to find out the context of a scene is the problem.

\begin{figure}[t]
        \centering
           \subfloat[Original image]{%
              \includegraphics[scale=0.45]{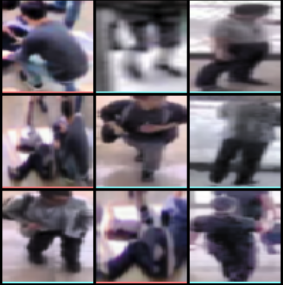}%
              \label{fig:left}%
           } 
           \qquad
           \subfloat[Unsupervised]{%
              \includegraphics[scale=0.45]{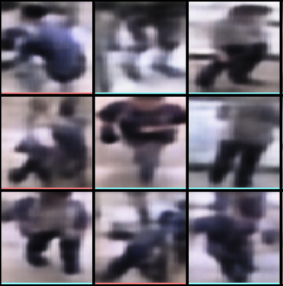}%
              \label{fig:middle}%
           }
           \qquad
           \subfloat[Negative learning]{%
              \includegraphics[scale=0.45]{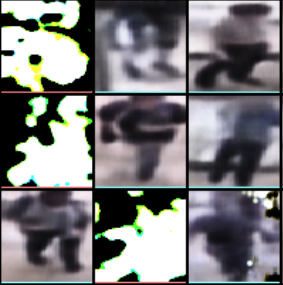}%
              \label{fig:right}%
           }
           \caption{Reconstruction of images(3x3) with unsupervised and negative learning methods. Colored line under each images shows the label of image. Orange is abnormal, green is normal data.}
           \label{fig:default}
    \end{figure} 
For abnormal detection, two types of model can be used, a discriminative model and a generative model.
Discriminative models are a class of models used in statistical classification, especially in supervised machine learning, which requires labeled data for training. 
On the other hand, the generative model doesn't need any labeled data for training, it can easily obtain desired data. 
For the training of the discriminative model, which requires labeled data, it is impossible to include all normal or abnormal events into the training data.
Because of its limitation to make a dictionary of normal and abnormal events or to define a standard for abnormal events, generating capability of the generative model is more preferred for abnormal detection.
With the generative model, by defining 'what is normal', the model can distinguish things far from normal as abnormal, which called novelty detection in other word. 

The generative model is trained only with ordinary situations, to learn a generalized feature of normal data.
Therefore, the model learns about distribution or a general feature of normal data to detect abnormals as outliers.,
However, there is another problem in this approach, which is the generalization capability of the generative model, that model can generate the anomalies too, which is not learned during the training and not desireable for our intension.

To solve this problem, limitation of reconstruction capability is suggested, which is called negative learning, in \cite{munawar2017limiting}.
Negative learning limits the reconstruction capability (LRC) of the generative model, and hinder the reconstruction of abnormal data by using labeled data.
Therefore, the network will only to generative learned data, and fails to generate unwanted feature.

In this work, we applied LRC for anomaly detection application.
For normal data, reconstruction is performed using the encoder-decoder structure as in the conventional generative model case.
But, for the abnormal data, negative learning \cite{munawar2017limiting} is applied.
The negative learning is supervised learning, which is learning a general distribution of normal data but also using labeled abnormal data during the training.
Negative learning can leverage the benefit of labeled abnormal data.
Because it's impossible to make a dictionary of every event, most anomaly detection methods use an unsupervised approach with generative models.
However, even if labeled data is expensive and very scares, it a waste if we don't use.
We can use abnormal data during the conventional training process, to leverage benefits from every data we got.
With negative learning, even a small amount of abnormal data can be used efficiently.

Conclusively this paper, we purpose an application for anomaly detection using the generative model with negative learning, which benefits from both labeled and unlabeled data, Showing limited reconstruction capability, only restrained to normal data.
\section{Related Works}

Labeless approach becomes a recent trend due to the nature of abnormal data, which is expensive to collect. Clustering\cite{leung2005unsupervised,zaheer2020claws,zaheer2020self,cleaning2020zaheer}, SVM network\cite{cortes1995support} and isolation forest\cite{liu2008isolation} are ways to perform anomaly detection without labels.
Because normal data is relatively easy and inexpensive to collect, training only with normal data become one branch, which called novelty detection.
Adversarially Learned One-Class Classifier\cite{sabokrou2018adversarially} and Old is Gold (OGNet) \cite{zaheer2020old} are GAN(Generative Adversarial Net)\cite{goodfellow2014generative} based architectures, which use discriminator for anomaly detection. During the training, the adversarial model trained on one-class, and inference time, generator and discriminator co-work for anomalous-classification. Deep-cascade net\cite{sabokrou2017deep} is also trained with normal data to learn the Gaussian model during the training and use this model for anomaly detection in each step of the deep network. 
These labeless, unsupervised learning approaches are more robust to an unknown anomaly compared to supervised learning.
However, they have an opposite problem of supervised learning approaches, which is hard to tell anomaly data near to the normal data, not as accurate as supervised learning in detecting known anomalies. OGNet attempts to specifically handle this issue by generating psuedo anomalies using an old state of the generator however, the overall training still utilizes only normal data.

Advance from labeless training, one intuition is to use weak-label for training. The idea proposed by Sultani\cite{sultani2018real}, and further extended in \cite{cleaning2020zaheer,zaheer2020claws,zaheer2020self}, used a video-level label, which only informs there is an anomaly in somewhere of video. With this weak video-level label, separate the anomaly video into small segments, assume at least one of those segments contain anomalous-scene for sure. Another study, Yang\cite{yang2015unsupervised} used implicit human support, from Youtube videos. They used user-edited short videos from youtube, assume that short videos(less than 4min) searched by specific keyword have a high chance to be consist of dominant scene correlated to the keyword.

Even if labeled anomaly data is expensive to collect, there is no reason not to use if we can have it. LRC\cite{munawar2017limiting} tried to leverage advantages from labeled anomaly data. By using labeled anomaly data, they limit the reconstruction capability of the generative model, which also capable of reconstructing anomaly data too. Similar to LRC, Bian\cite{bian2019novel} also used negative learning stage to limit the reconstruction of GAN. They trained on normal data, but also get advantage from the labels.

\section{Method}

In this section, methodology and objective are defined.
First, assume that two data distributions X and Y are in the same signal space.
And x, y be the instances that represent each distribution, so that X and Y can be expressed as $X = \{x_1,x_2,x_3,...,x_k\}$, $Y = \{y_1,y_2,y_3,...,y_j\}$.
Our goal is to make the network regenerate X effectively while failing for Y. Here X and Y represent the normal and abnormal data.
Let the distribution of regenerated X be $\hat{X} $ and the distribution of regenerated Y be $\hat{Y}$.
Then the goal is to maximize the following equation:
\begin{equation}
    p_\theta(\hat{X}|X)-p_\theta(\hat{Y}|Y)= \sum_{i}^{k}\log_\theta{(\hat{X}|X)}- \sum_{i}^{j}\log_\theta{(\hat{Y}|Y)}
\end{equation}
that network will reconstruct very well the from the X, but fail to reconstruct from Y.

For simplicity, the mean square error is used for loss function. 
Network parameter $\theta = \{\alpha,\beta\}$ is being trained, which $\alpha$ is parameter of encoder and $\beta$ is parameter for decoder.
For the X, the normal data, the objective is to minimize the loss to find optimal $\theta^*$:
\begin{equation}\label{positive_learning}
    \theta^* = \underset{\theta}{argmin}\sum_{i=1}^{k}\{\hat{x_i}-x_i\}^2
\end{equation}
the theta that can well reconstruct from normal distribution.
On the other hand for the Y, the abnormal data, the objective is the opposite of that, to maximize the loss to hinder the reconstruction of the network:
\begin{equation}\label{negative_learning}
    \theta^* = \underset{\theta}{argmax}\sum_{i=1}^{k}\{\hat{y_i}-y_i\}^2
\end{equation}
Negative-learning is the opposite concept of normal(positive) learning, which makes network hard to reconstruct undesired data.
Both equations are trying to optimize the same parameter, $\theta$ simultaneously.

The algorithm \ref{alg:rmleft} shows the process of training with negative learning. 
For negative learning, everything is identical to the conventional training sequence, only the loss for the abnormal data is changing.
For normal data represented as X, Eq.\eqref{positive_learning} is used.
However, for the abnormal data Y, Eq.\eqref{negative_learning} is used, which mean negative learning.
Two loss functions are working with the same parameter, to achieve different goal respectively, generate good normal reconstruction and fail to do the same thing for anomaly data.
And because there are unbalance in the data quantity, in LRC\cite{munawar2017limiting} used multiple iterations for negative learning to keep the balance between normal and abnormal data.
For our case, because the ratio of normal and abnormal data was 10:1, we tried an experiment for the weighted loss.
Instead of multiple iterations, extra 10-time weight is given for the abnormal data during the negative learning in the experiment.
However, there was no significant difference between weighting and without-weighting in the test results, so we decide to use the same equation with only the opposite sign for positive and negative learning.

\begin{algorithm}[t]
\caption{}\label{alg:rmleft}
\begin{algorithmic}[1]
\Require
X: Normal dataset
Y: Abnormal dataset
\State $TRAINING\_START$ 
\While{NOT TERMINATION CONDITION}{
\If{$DATA \in X$} {DO POSITIVE LEARNING} 
\ElsIf{$DATA \in Y$} {DO NEGATIVE LEARNING} 
\EndIf}
\EndWhile
\State $TRAINING\_END$
\end{algorithmic}
\end{algorithm}

\section{Experiments}


\subsection{Data set}

The experiment is separated into two phases.
First train and test on a new dataset, which is our own, and next test on the public MNIST dataset.
The first dataset, which named `lobby dataset', is consist of records from a surveillance camera installed on the entrance of the building.
The video contains people entering and leaving the building through the main gate.
We collected 20 normal videos and each of them is 24 hours long.
Abnormal video is 2 hours long and made of actors playing certain situations, like sudden fall or running, an unusual gathering or running of people.
With the videos, we detect the bounding box of appearing human in the scenes and extract the cropped human-size images.
We collected 55,247 normal image patches and 3,569 abnormal image patches.
Then divide them into $7:1:2=train:validation:test$ for experiment.
By including the same sequence to the same data group, a similar scene doesn't appear in train and test dataset at the same time during the training. That is mot all the anomaly scene contained in train dataset, so the network will face an unknown anomaly in the test step.

\subsection{Network Structure}

We use the Adversarial Auto-Encoder \cite{makhzani2015adversarial} as a basic structure and for some part of the network refer to the DCGAN \cite{radford2015dcgan} for better performance.
However, any Auto-encoder structure can be used instead. 
Images are transformed into 64x64 size RGB 3 channel and fed to network as input.
Every training was done with 1000 epoch with batch size 1024. 
Adam is used for an optimizer and we used the same parameter as paper\cite{kingma2014adam}.
Batch normalization\cite{ioffe2015batch} is applied after convolution layer and we used ReLU as an activation function for both encoder and decoder. 
However, only the last layer of the decoder, we used tanh exceptionally\cite{radford2015dcgan}.
With adversarial part of AAE, we simply impose gaussian distribution to the latent code\cite{makhzani2015adversarial} for simplicity.



\subsection{Comparison methods}
With the same structure and dataset, only the loss function is changed for comparison.
First, the unsupervised setting has experimented as a baseline.
During the training, only normal train set was used, which is identical to the train set used for the negative learning phase.
The unsupervised network was trained to learn only normal situations and tested on the dataset, that contains both normal and abnormal scene, to distinguish them.
For the supervised negative learning approach, the network learned normal data in same manners of the unsupervised model.
However, after training with normal data, the supervised model trained with extra abnormal data set by negative learning.
Both unsupervised and supervised approach used the same data set (train \& test), but extra abnormal data was used for negative learning. This extra abnormal train data set was excluded during the unsupervised phase.
This separation is to show the effect of leveraging the advantage of existing anomaly data.



\begin{figure}
\centering
   \subfloat[Unsupervised]{%
      \includegraphics[scale=0.28]{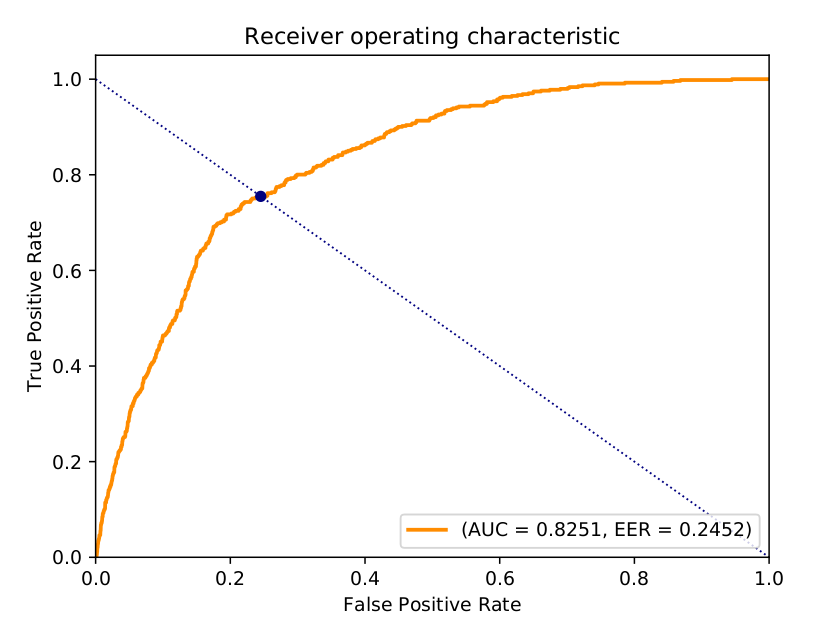}%
      \label{fig:left}%
   } 
   \subfloat[Negative learning]{%
      \includegraphics[scale=0.28]{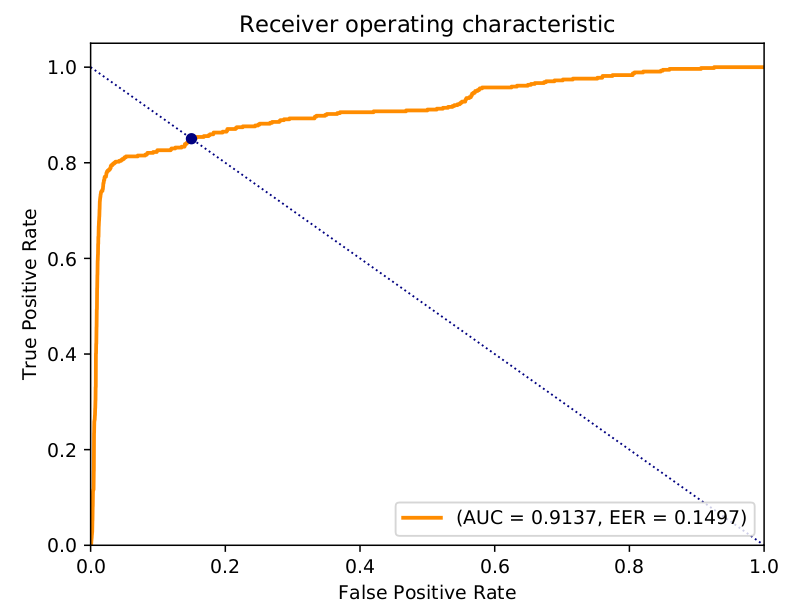}%
      \label{fig:middle}%
   }
   \caption{ROC comparision of unsupervised and negative learning.}
     \label{fig:ROC}
\end{figure}

\begin{figure*}
\centering
  \includegraphics[width=\textwidth]{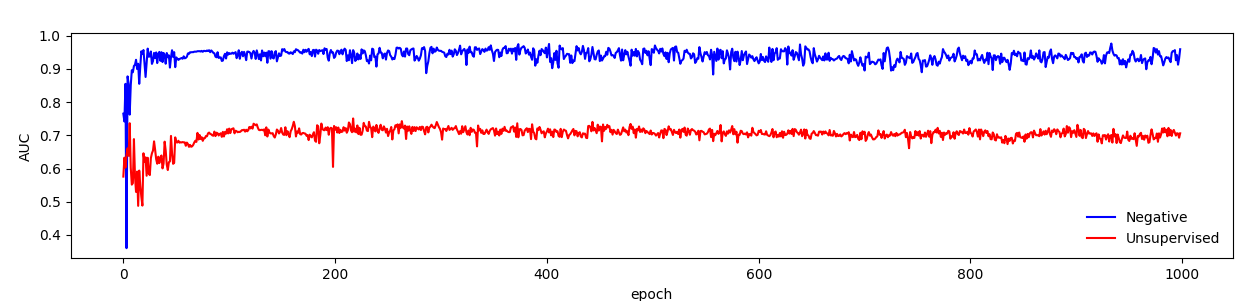}
  \caption{Negative learning AUC change during the training.}
  \label{fig:AUC}
 \end{figure*}
\begin{figure*}
  \includegraphics[width=\textwidth]{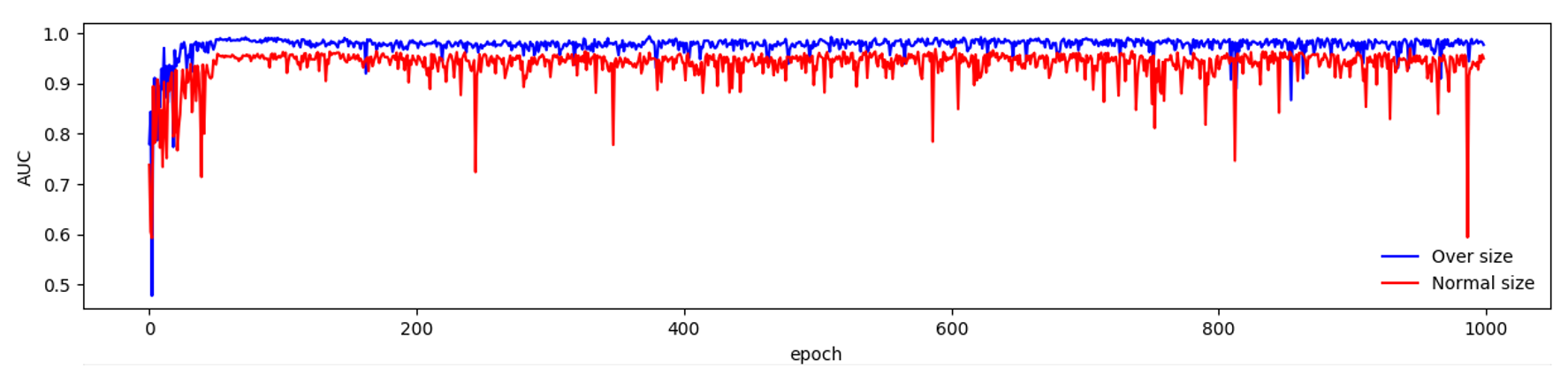}
  \caption{Oversize learning AUC change during the training.}
  \label{fig:AUC_over}
\end{figure*}

\begin{figure}
  \centering
    \includegraphics[width=0.8\linewidth]{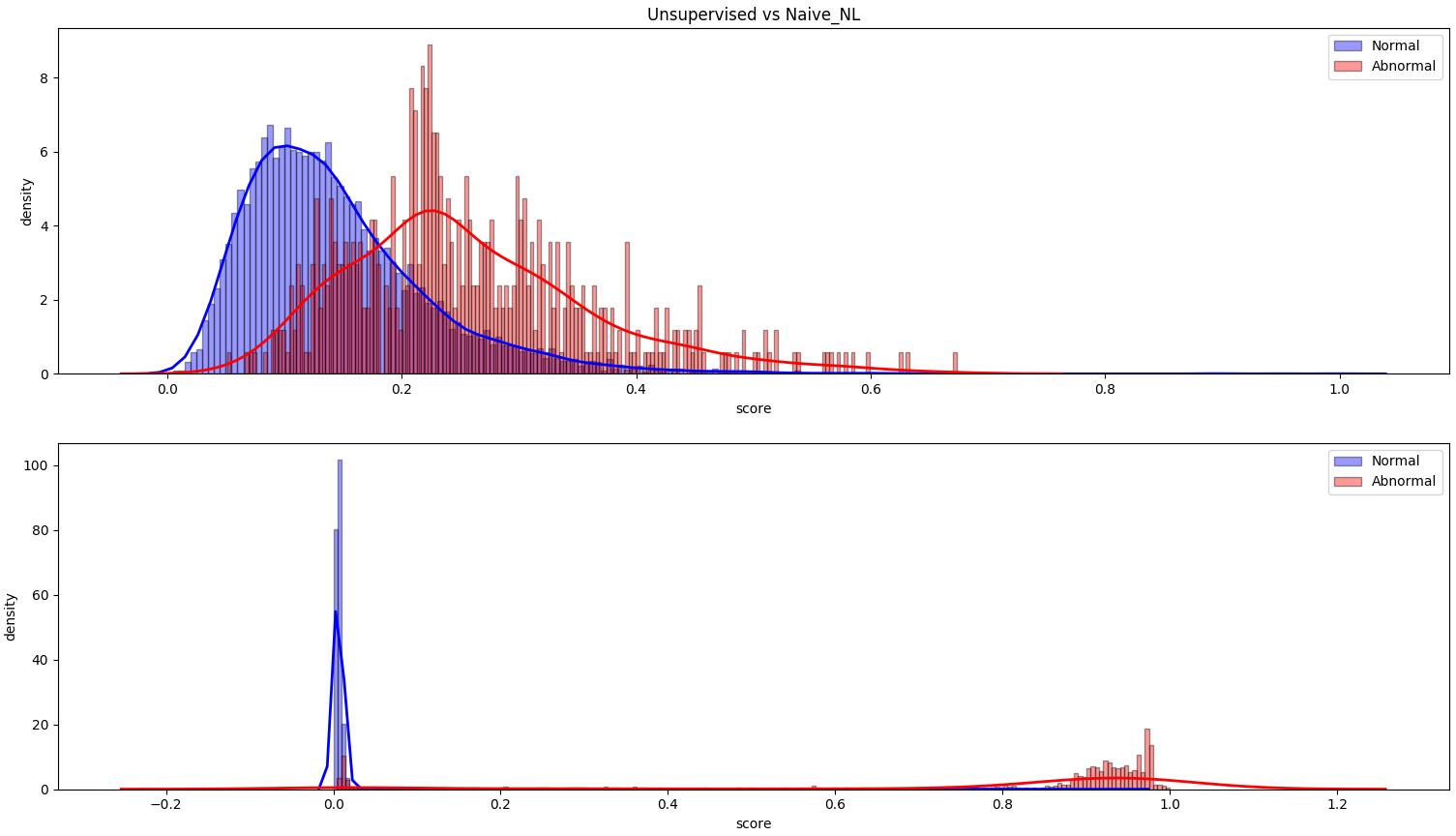}
  \caption{Score distribution after unsupervised and negative learning}
  \label{fig:dist}
\end{figure}

Receiver Operating Characteristic (ROC) and Area Under Curve (AUC) are calculated for evaluation, during the training(Fig. \ref{fig:AUC}) and after(Fig.\ref{fig:ROC}) training with validation set and test set respectively.
The result during the training shows that AUC of the negative learning exceeded that of unsupervised learning for all times(Fig. \ref{fig:AUC}). 
Negative learning method's AUC value get close to around 0.95, which outperformed the AUC value of unsupervised learning, stays around 0.73.    
Overall performance and discriminating capability of negative learning model were much better than the unsupervised model, which is only trained with normal data. 

At the test time, the AUC value of Unsupervised model rises to 0.82. 
Despite the AUC value of negative learning model slightly decreased, compared to that result during training, but still very high compared to the unsupervised model, around 0.91.
The training was done eight times each for both methods and we selected the best model for evaluation.
The result of all training is shown in Table. \ref{tab:auc_comp}.
Negative learning performs better given average AUC value, 0.89 compared to 0.76 of unsupervised learning.
Also, the standard deviation of AUC value shows that negative learning is more stable in training.

The Fig. \ref{fig:dist} shows the Score distribution of the dataset
The score calculation is based on reconstruction error, which bigger the reconstruction error leads to higher the score.
After computing the score, scores are scaled to 0-1.
Upper one in the Fig. \ref{fig:dist} is the unsupervised method and the other is the negative learning method.
In each distribution, blue represents normal data and red represents abnormal data.
The upper distribution in Fig. \ref{fig:dist} is the unsupervised method. 
There is a large overlap between normal and abnormal data.
That means it's hard to tell normal from abnormal with this model.
On the other hand, in the below distribution in Fig. \ref{fig:dist}, two data sets(normal and abnormal) have a large gap between them, so the network can easily detect anomalies.

\begin{figure}[htbp]
    \centering
    \subfloat[Unsupervised 0 original]{{\includegraphics[width=0.22\textwidth]{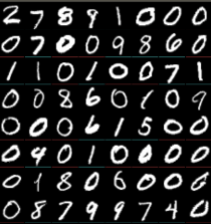}}}
    \hfill
    \subfloat[Unsupervised 0 reconstructed]{{\includegraphics[width=0.22\textwidth]{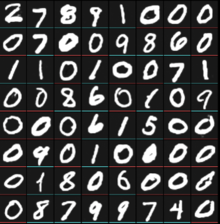}}}
    \hfill
    \subfloat[Unsupervised 1 original]{{\includegraphics[width=0.22\textwidth]{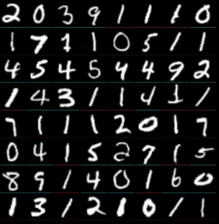}}}
    \hfill
    \subfloat[Unsupervised 1 reconstructed]{{\includegraphics[width=0.22\textwidth]{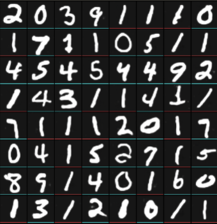}}}

    \subfloat[NL 0 original]{{\includegraphics[width=0.22\textwidth]{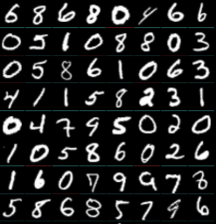}}}
    \hfill
    \subfloat[NL 0 reconstructed]{{\includegraphics[width=0.22\textwidth]{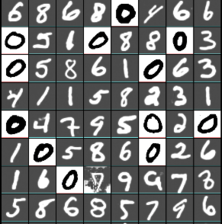}}}
    \hfill
    \subfloat[NL 1 original]{{\includegraphics[width=0.22\textwidth]{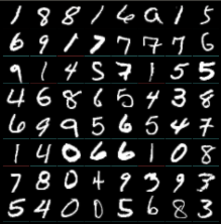}}}
    \hfill
    \subfloat[NL 1 reconstructed]{{\includegraphics[width=0.22\textwidth]{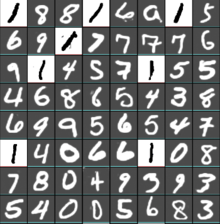}}}    
    \caption{Result of MNIST reconstruction Comparison between unsupervised and negative learning(NL) methods. Digit 0 and 1 are used as anomaly data. The network was trained with MNIST data but without anomaly number(0 \& 1). Unsupervised network well generalized the data and reconstructed anomaly very good. NL method also reconstructed well but anomaly is distinguishable from others. }
    \label{MNIST_output}
\end{figure}


\begin{table*}[]
\resizebox{\textwidth}{!}{
\begin{tabular}{llllllllllll}
Method $\|$ Anomaly Number & 0     & 1     & 2     & 3     & 4     & 5     & 6     & 7     & 8     & 9     & Avg.  \\ \hline
Unsupervised (Baseline) & 73.57 & 17.06 & 62.54 & 55.48 & 43.33 & 45.16 & 56.41 & 41.81 & 58.63 & 42.91 & 49.69 \\
Negative Learning       & 99.64 & 99.59 & 98.99 & 99.40 & 99.34 & 99.11 & 99.14 & 98.82 & 98.86 & 98.17 & 99.11
\end{tabular}
}
\caption{AUC comparison with MNIST dataset. Anomaly numbers were used as anomaly input in each test.}
\label{tab:auc_mnist}
\end{table*}

\begin{table*}[]
\resizebox{\textwidth}{!}{%
\begin{tabular}{lllllllllll}
Method $\|$ Test seed   & 1     & 2     & 3     & 4     & 5     & 6     & 7     & 8     & Avg.  & STDEV \\ \hline
Unsupervised (Baseline) & 82.70 & 82.51 & 66.45 & 80.39 & 63.86 & 80.08 & 81.28 & 70.70 & 75.99 & 0.077 \\
Negative Learning       & 92.66 & 91.37 & 87.45 & 83.24 & 89.47 & 88.66 & 90.68 & 88.61 & 89.01 & 0.028
\end{tabular}%
}
\caption{AUC comparison between negative learning and unsupervised learning. test was done 8 times with 1000 epoch each.}
\label{tab:auc_comp}
\end{table*}

\begin{figure}
    \centering
    \includegraphics[scale=0.2]{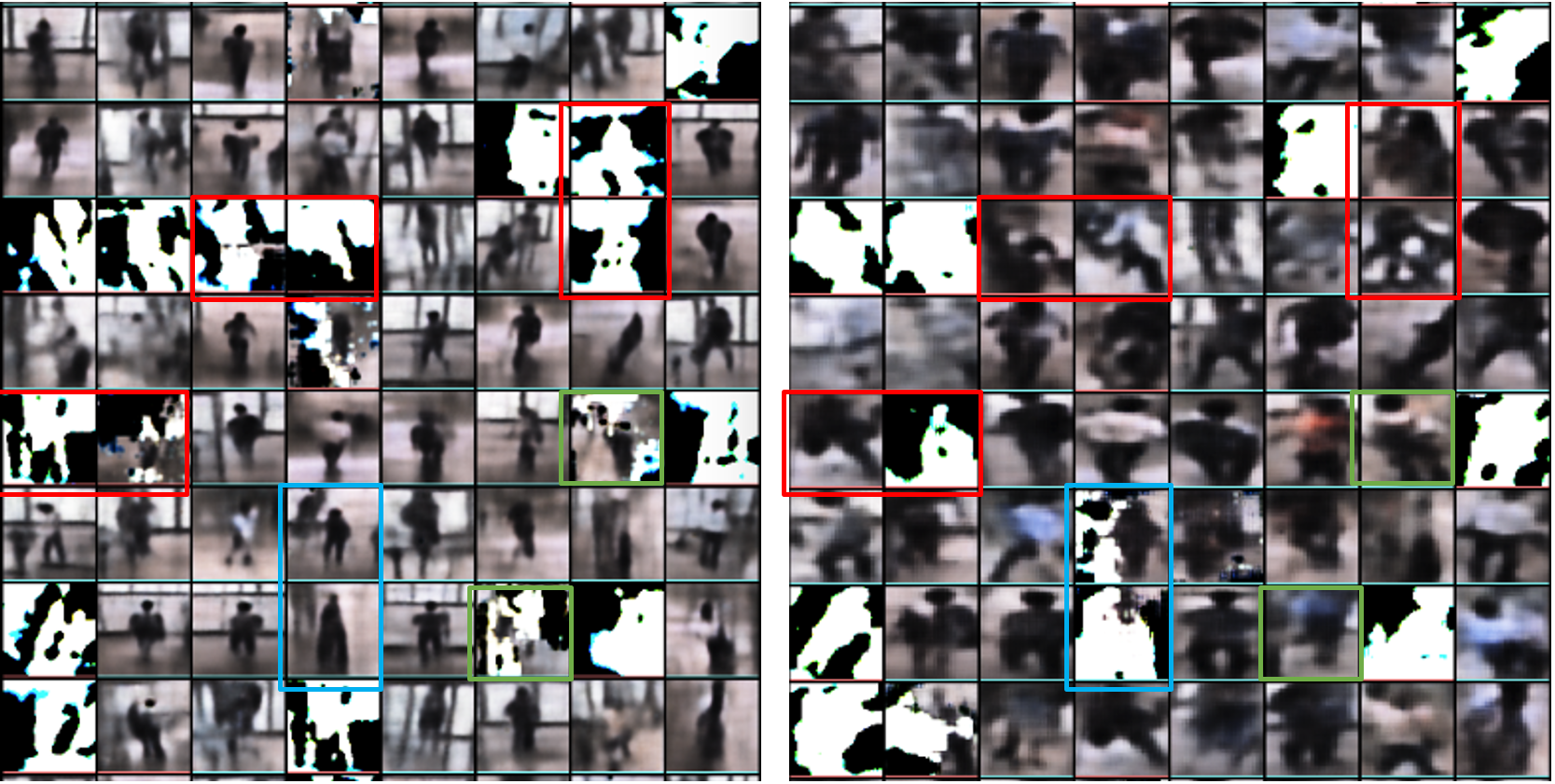}
    \caption{comparison between over crop size test and normal crop size test. Green or orange line under each image show data is normal(green) or abnormal(orange). Red box indicates abnormal images discriminated correctly in oversize, but not in normal size. Blue box indicates normal images predicted as abnormal in normal size training, but discriminated properly as normal in oversize test. Finally green box is normal images which is properly discriminated in normal size test but failed to predict in oversize test.}
    \label{fig:over_n_compare}
\end{figure}

\subsection{Test on MNIST dataset}
In the second phase of experiment, network tested on MNIST, public dataset.
For the training, first set one digit from 0-9 as anomaly number.
Then train network with the other numbers.
The MNIST dataset was separated into 8 : 2 = Train : Test.
After training 100 epoch with 256 batch size, the test is done by ten times with all numbers(0-9) as anomaly number at each test case.
The Table. \ref{tab:auc_mnist} show the experimental result.
With unsupervised learning, few numbers were barely discriminated, compared to average 99 of negative learning result.

The Fig. \ref{MNIST_output} show outputs of the MNIST experiment.
Four experiment results are shown in Fig. \ref{MNIST_output}. 
Unsupervised learning reconstructs abnormal number very well, even they never trained with them, because of their generalization capability.
Negative learning also reconstructed well, but in an inverted pixel, which makes network easily separate them.
The average AUC of negative learning was 0.99, outperformed unsupervised learning 0.49(Table. \ref{tab:auc_mnist}), which shows the effect of negative learning in anomaly detection.


\subsection{Oversize test}

After verification in public dataset, an extra experiment was performed, with an oversized crop image, assuming that bigger image will contain more information to help the prediction of the network.
The Oversize crop image size is 2.5 times bigger in width and 1.5 time in the height of the object bounding box.
However, before fed to the network, also resized into 64x64, same like normal crop images used in the previous experiment.
Other things(architecture, loss function, dataset) are all the same condition. Only the initial crop size is bigger between them.
AUC during the training is shown in Fig. \ref{fig:AUC_over}.
Oversize training converges faster and higher than normal size training.
Table. \ref{Over_table} shows the comparison of the test result for oversize between normal negative learning.
The average AUC of oversize training is 0.04 higher compared to normal negative training result.
The Fig. \ref{fig:over_n_compare} show more detailed output of test.
Oversize training reduced true negative(normal predicted as abnormal) and false positive(abnormal predicted as normal) cases.
With this test, we confirm the effect of bigger crop size, which contains more appearance information of the image, does help network.

\begin{table}[]
\centering
\begin{tabular}{cccccccc}
\multicolumn{1}{l}{\#} & \multicolumn{1}{l}{1} & \multicolumn{1}{l}{2} & \multicolumn{1}{l}{3} & \multicolumn{1}{l}{4} & \multicolumn{1}{l}{5} & \multicolumn{1}{l}{6} & \multicolumn{1}{l}{avg} \\ \hline
Oversize               & 0.9850                & 0.9808                & 0.9759                & 0.9775                & 0.9757                & 0.9804                & 0.9792                  \\ \hline
Normal size            & 0.9379                & 0.9164                & 0.9352                & 0.9317                & 0.9227                & 0.9367                & 0.9301                 
\end{tabular}
\caption{Test AUC result on Oversize crop image.}
\label{Over_table}
\end{table}

\subsection{Scaling Negative loss} 
There is some work called Auto-encoding Binary Classifier(ABC) \cite{ABC2019abc}, archived but yet published, which also utilize LRC for their work.
The interesting point of this work is limiting the boundary of negative loss.
Without limitation, the negative loss can be really huge so that can may overwhelm the positive loss.
Similar to this work, We used a different loss to scale negative loss during the training:
\begin{equation}
    loss = \sum^{j}_{i}\exp^{-(\hat{y}_i-y_i)^2}
\end{equation}
This will scale the range of negative loss between 0 to 1, which is maximum reconstruction negative loss will be 1, and become smaller as reconstruction loss gets bigger, that network will aim to minimize this loss.
With this loss, we compare the result to the result of naive loss, which is Eq.\ref{negative_learning}.
Detail of comparison is shown in Table. \ref{ABC_table}.
Average AUC of both tests was very similar, or we can say limitation doesn't affect the result.
Further, we analyze the output of both tests in detail, which is explained in the next section.

\begin{table}[]
\centering
\begin{tabular}{cccccccccc}
\multicolumn{1}{l}{Methods} & \multicolumn{1}{l}{1} & \multicolumn{1}{l}{2} & \multicolumn{1}{l}{3} & \multicolumn{1}{l}{4} & \multicolumn{1}{l}{5} & \multicolumn{1}{l}{6} & \multicolumn{1}{l}{7} & \multicolumn{1}{l}{8} & \multicolumn{1}{l}{avg} \\ \hline
Scale                       & 0.8621                & 0.8869                & 0.8954                & 0.8752                & 0.9056                & 0.9049                & 0.8830                & 0.9221                & 0.8919                  \\ \hline
Naive                       & 0.9266                & 0.9137                & 0.8745                & 0.8324                & 0.8947                & 0.8866                & 0.9068                & 0.8861                & 0.8901                 
\end{tabular}
\caption{Test AUC result on Scaled negative loss test.}
\label{ABC_table}
\end{table}

\subsection{Score Tracking}

\begin{figure*}
\centering
    \subfloat[Naive abnormal]{{\includegraphics[width=0.44\textwidth]{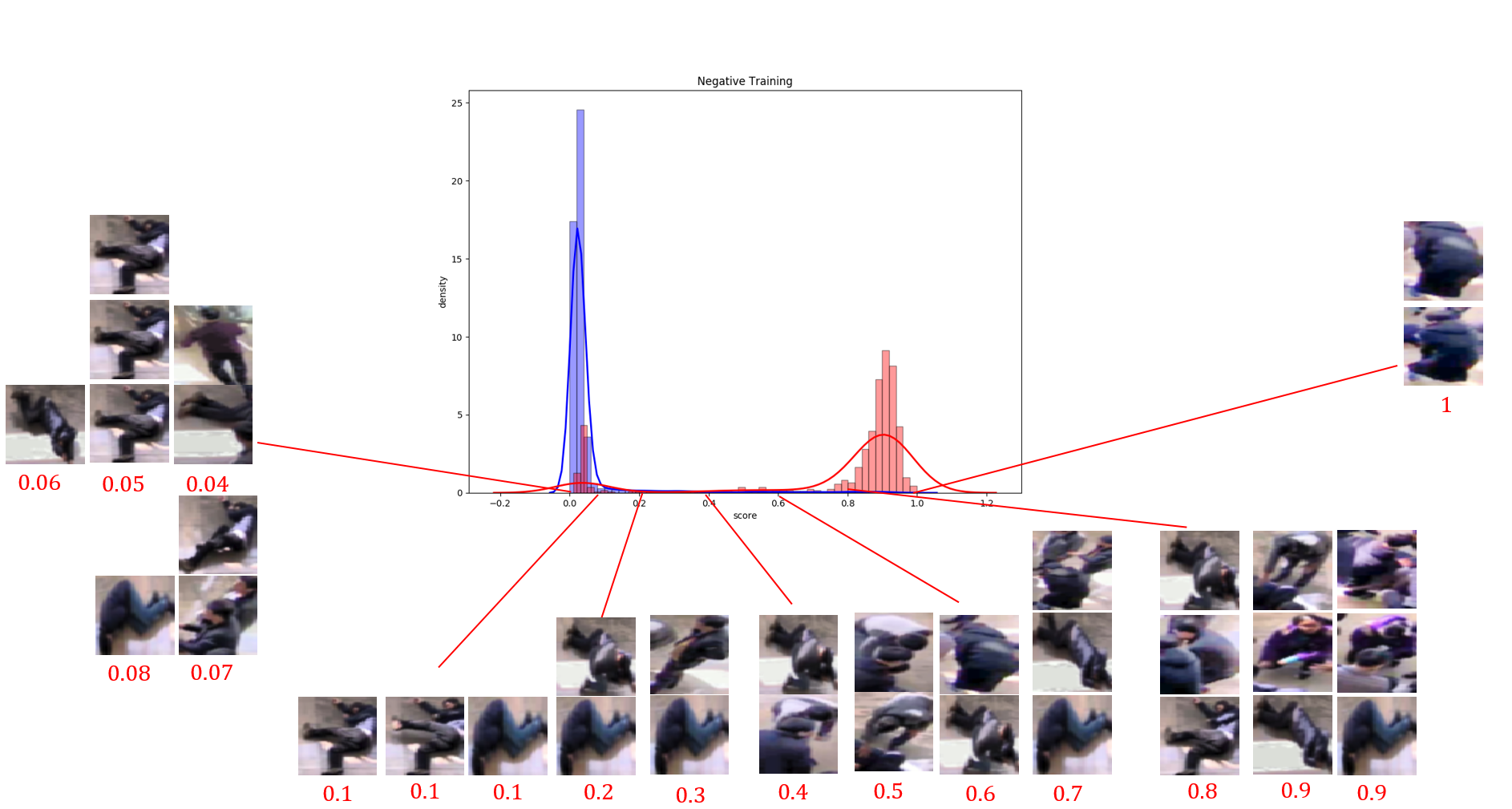}}}
    \hfill
    \subfloat[Naive normal]{{\includegraphics[width=0.44\textwidth]{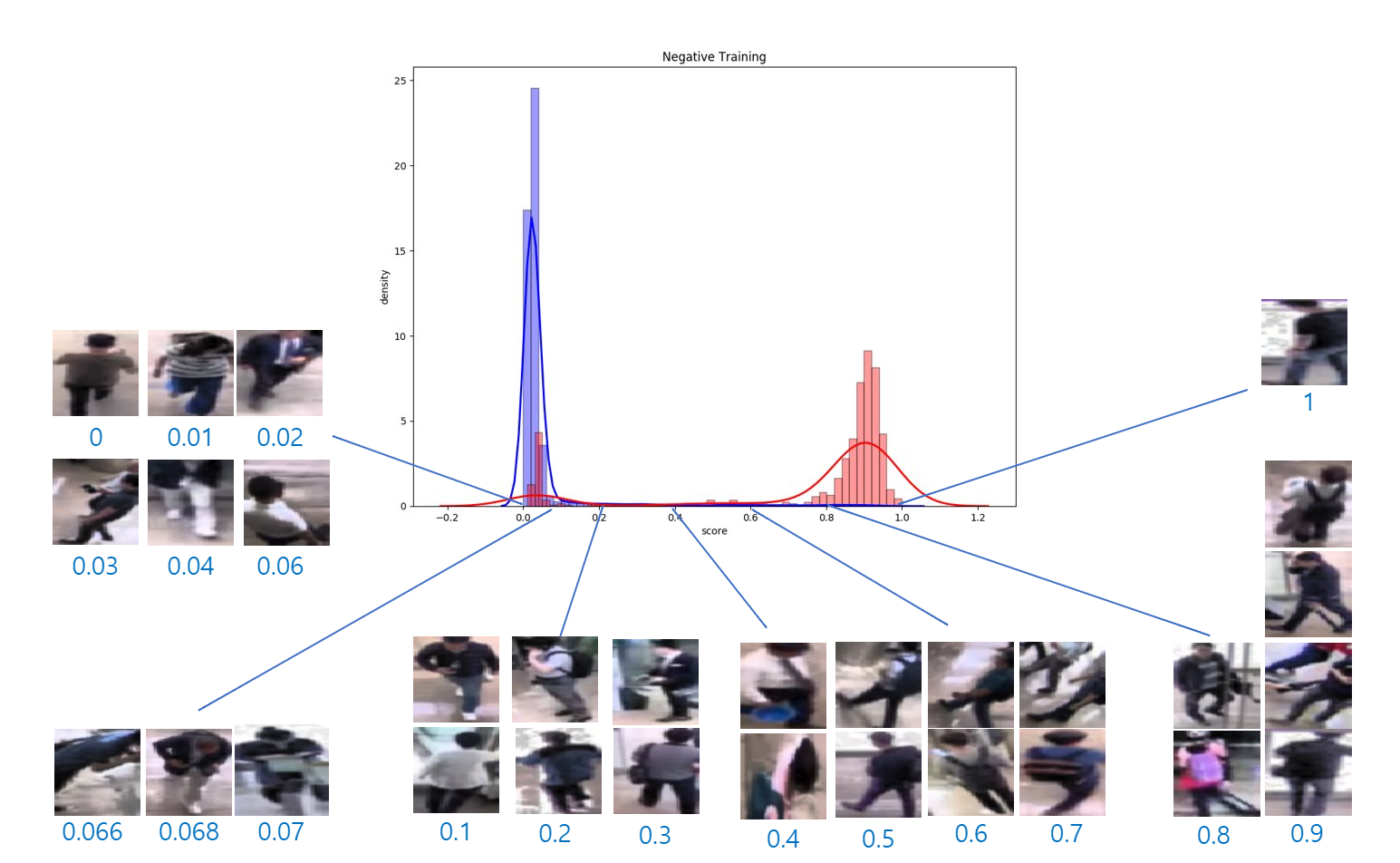}}}
    \hfill
    \subfloat[Scale abnormal]{{\includegraphics[width=0.44\textwidth]{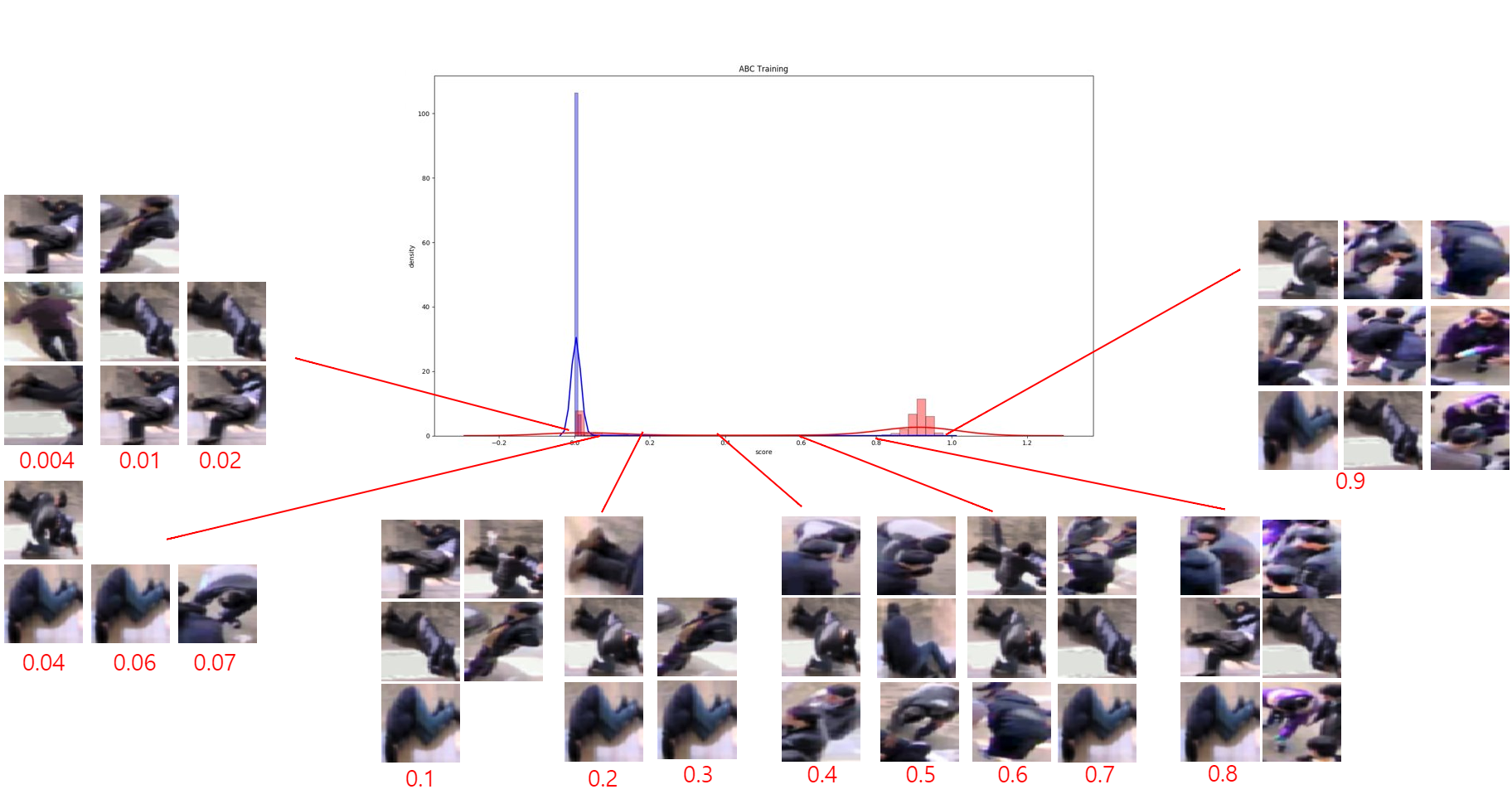}}}
    \hfill
    \subfloat[Scale normal]{{\includegraphics[width=0.44\textwidth]{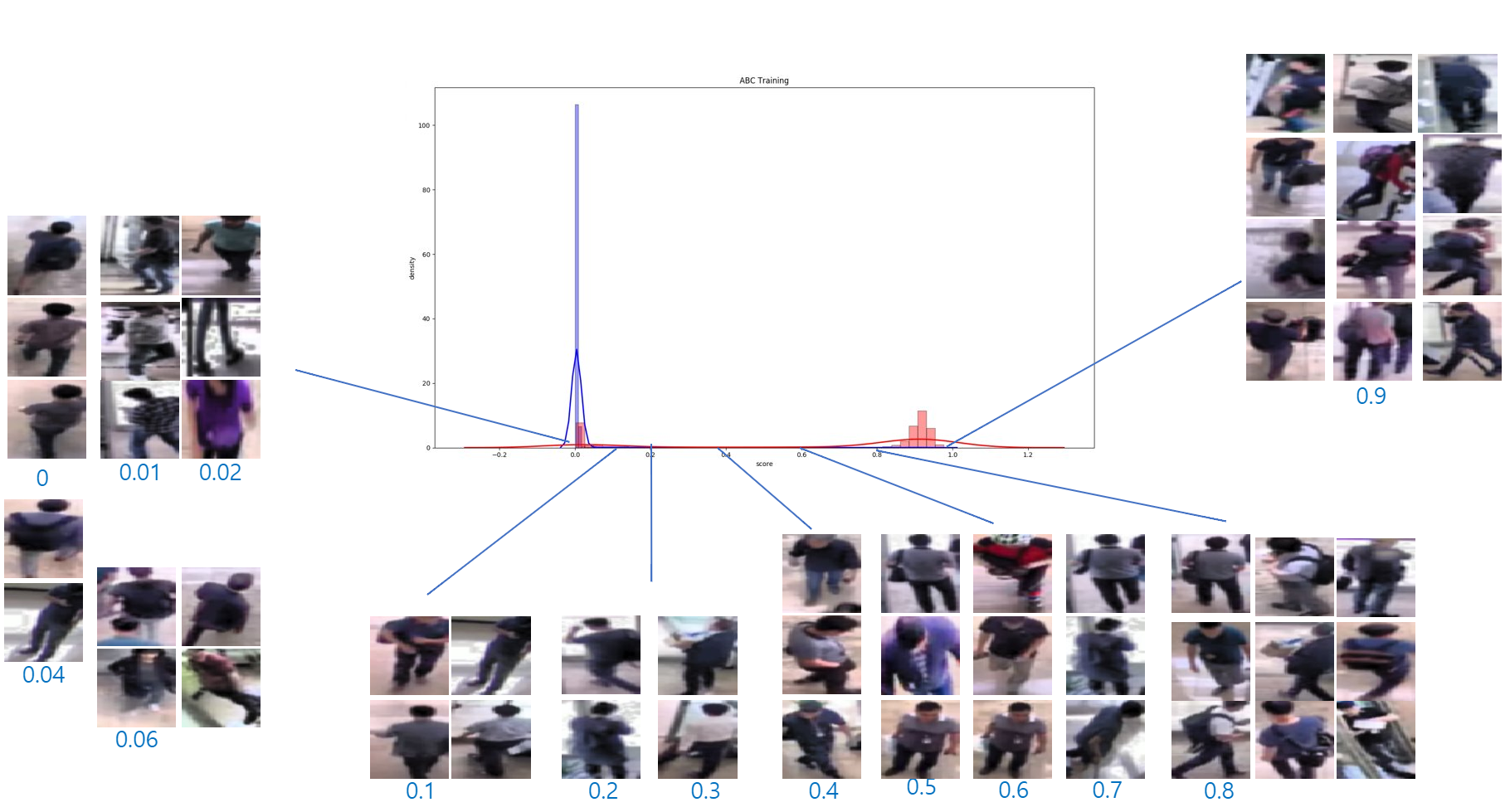}}}
    \caption{Result of Score tracking.}
    \label{score_track}
\end{figure*}

In this section, we track the score distribution of naive and ABC network, to see the effect of each approach.
Fig. \ref{score_track} show the distribution of images in detail.
The score is determined by the loss, scaled into 0-1.
(a),(b) shows the distribution of abnormal and normal images of naive training and (c), (d) shows the distribution of Scaled training.
Looking at the (b),(d) we can find out how the network gives the score.
In general, someone wears a colored shirt or bag gets a high score.
Also, when they show a big angle in their leg, which mean walking horizontally against the camera got the high score. This phenomenon reflects the feature of the lobby.
On the other hand, looking at the abnormal image score distribution, there are some peculiar parts in their appearance.
The general part follows the rule like normal distribution(color, angle), but some images are shown in every point of the score distribution.
We figured out that with only 2d appearance without temporal information, it is difficult to tell whether you are standing or lying.
Even if we consider it, it is difficult to understand the causes seen in all sections.
This seems to be the limit of this network that uses only instantaneous images as input.

\section{Conclusion}
We proposed an application of anomaly detection utilizing negative learning to limit the reconstruction capability of the generative model.
Supervised learning with negative loss help network gets advantage from labeled anomaly data.
The effect of using labeled data, along with conventional unsupervised manner, was confirmed with outperforming result of experiments.
Still, there are few more work to do, like the current network only receives frame as input which doesn't contain any temporal information.
With only the appearance information, it is hard to tell some situations shown in section 4-G.
Better performance is expected with adding temporal information, which is left for future work.

\subsubsection*{Acknowledgements}
This  work  was  supported  by  the  ICT  R\&D  program  of  MSIT/IITP. [2019-0-01309,  Development  of  AI  Technology for  Guidance  of  a Mobile  Robot  to  its  Goal  with  Uncertain  Maps  in  Indoor/Outdoor Environments]

%
%
%
%

{\small
\bibliographystyle{unsrt}
\bibliography{samplepaper}
}

\end{document}